\documentclass{article}
\usepackage{spconf,amsmath,graphicx}

\usepackage{amssymb}

\usepackage{booktabs}
\usepackage{subcaption}

\let\OLDthebibliography\thebibliography
\renewcommand\thebibliography[1]{
  \OLDthebibliography{#1}
  \setlength{\parskip}{0pt}
  \setlength{\itemsep}{0pt plus 0.3ex}
}

\def\L{{\cal L}}

\title{Direct Handheld Burst Imaging to Simulated Defocus}
\name{Meng-Lin Wu \qquad Venkata Ravi Kiran Dayana \qquad Hau Hwang}

\address{Qualcomm Technologies, Inc.}
\begin{document}
%


\maketitle
\begin{abstract}
A shallow depth-of-field image keeps the subject in focus, and the foreground and background contexts blurred. This effect requires much larger lens apertures than those of smartphone cameras. Conventional methods acquire RGB-D images and blur image regions based on their depth. However, this approach is not suitable for reflective or transparent surfaces, or finely detailed object silhouettes, where the depth value is inaccurate or ambiguous.

We present a learning-based method to synthesize the defocus blur in shallow depth-of-field images from handheld bursts acquired with a single small aperture lens. Our deep learning model directly produces the shallow depth-of-field image, avoiding explicit depth-based blurring. The simulated aperture diameter equals the camera translation during burst acquisition. Our method does not suffer from artifacts due to inaccurate or ambiguous depth estimation, and it is well-suited to portrait photography.


\end{abstract}

\section{Introduction}
\label{sec:intro}

\begin{figure*}[t]
\centering
    \begin{subfigure}[b]{0.2285\textwidth}
        \includegraphics[width=\textwidth]{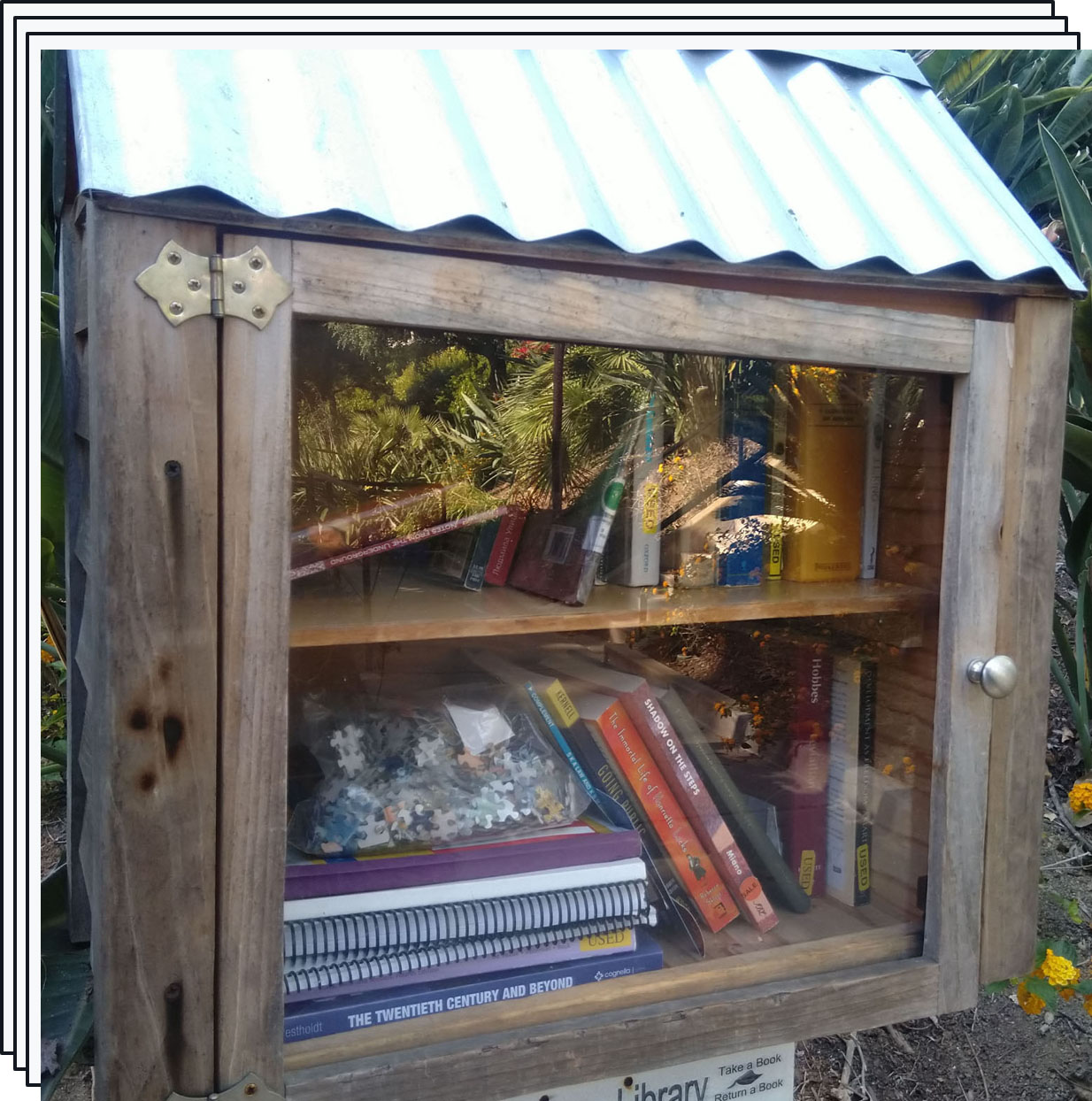}
        \caption{Input handheld burst}
    \end{subfigure}
    \begin{subfigure}[b]{0.22\textwidth}
        \includegraphics[width=\textwidth]{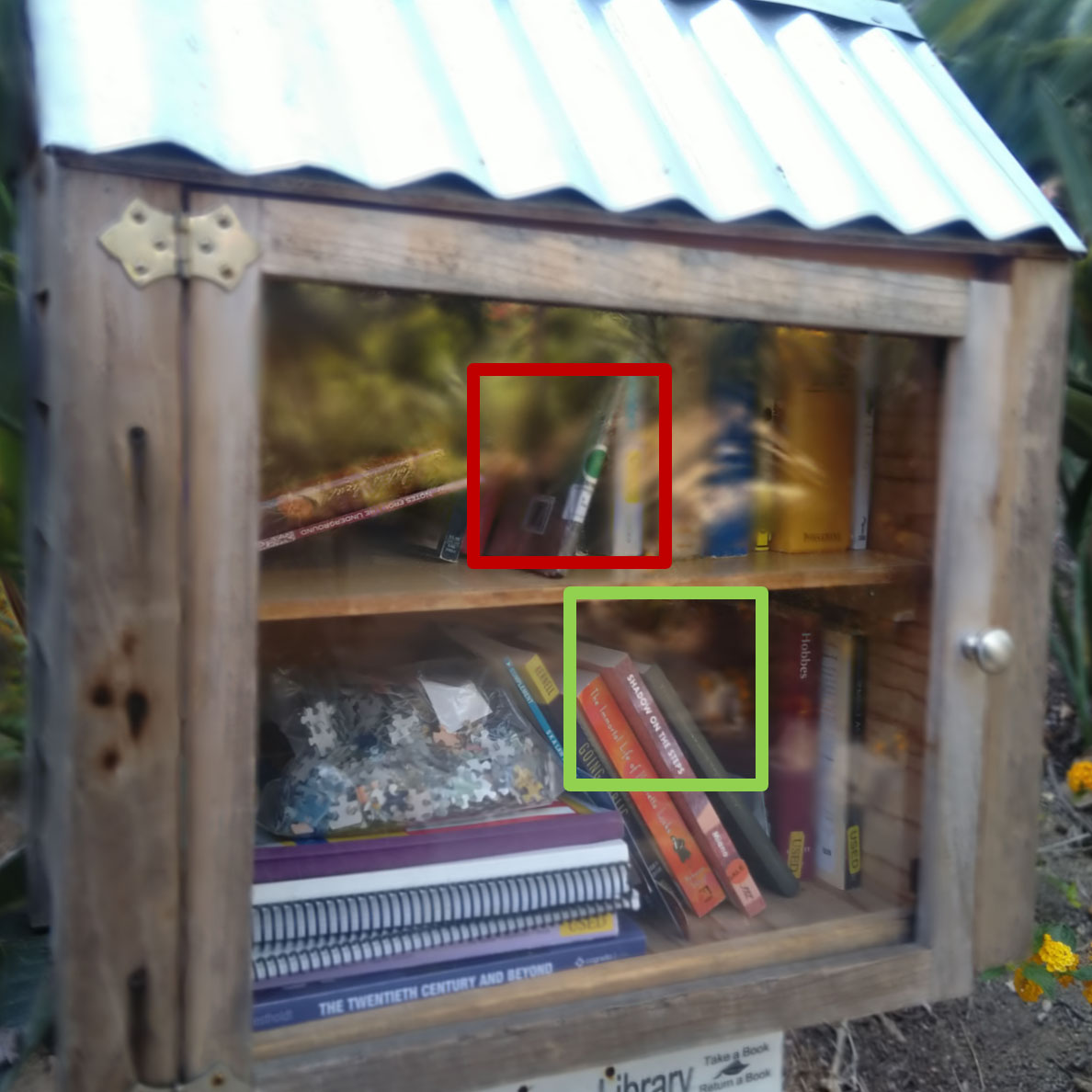}
        \caption{Refocusing on books}
    \end{subfigure}
    \begin{subfigure}[b]{0.22\textwidth}
        \includegraphics[width=\textwidth]{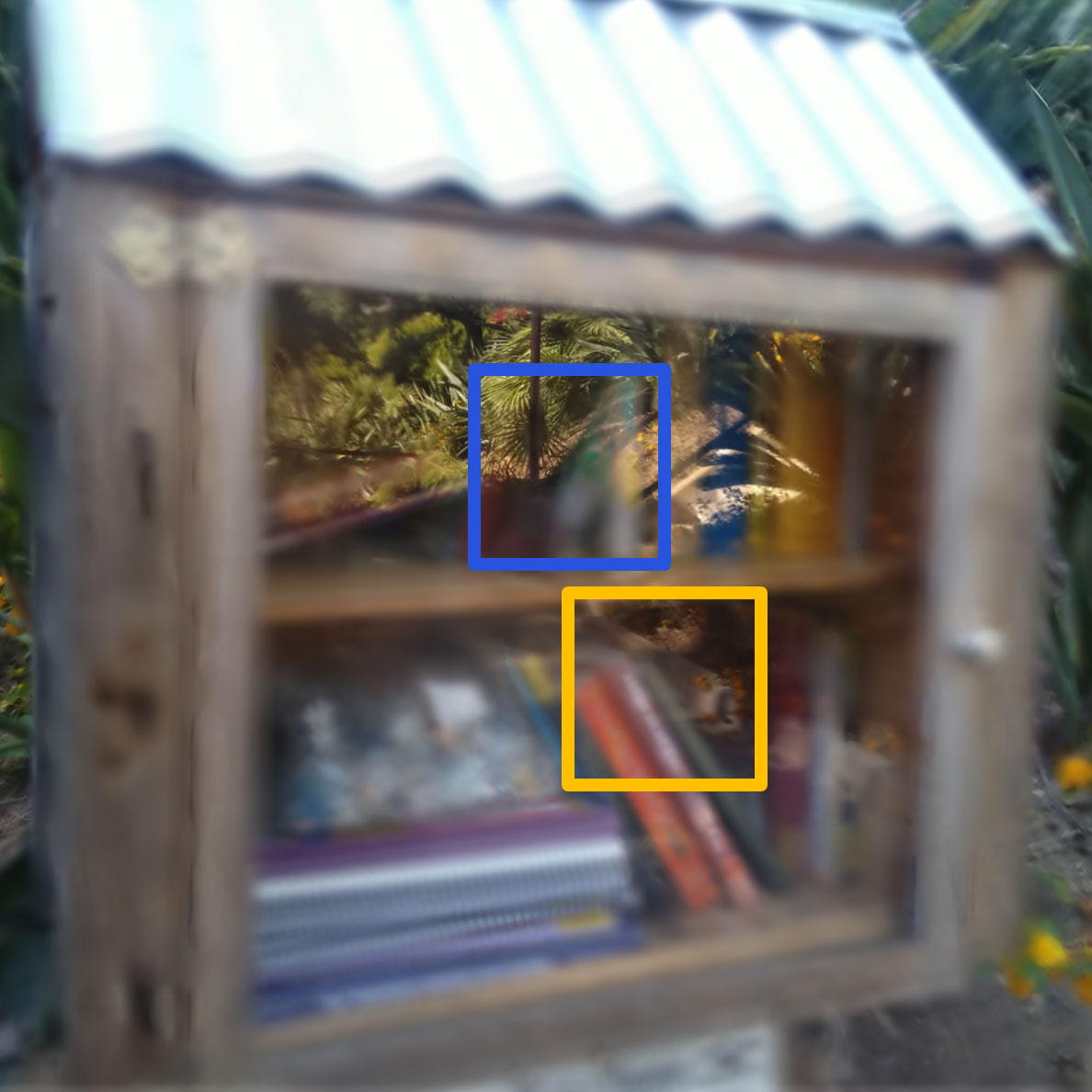}
        \caption{Refocusing on reflection}
    \end{subfigure}
    \begin{subfigure}[b]{0.22\textwidth}
        \includegraphics[width=\textwidth]{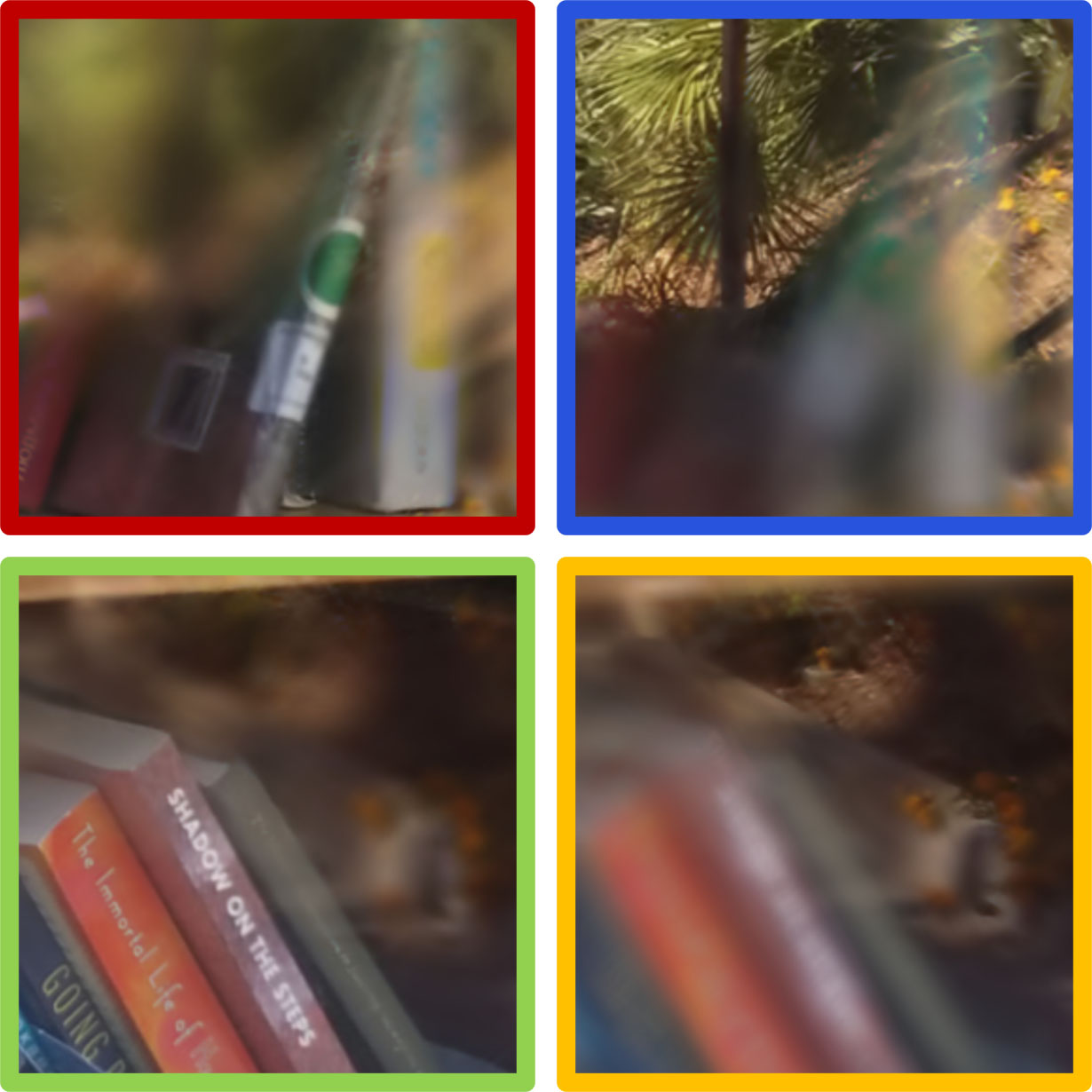}
        \caption{Magnified view}
    \end{subfigure}
    \vspace{-0.2cm}
	\caption{Our method takes an input handheld burst and directly generates the shallow DOF output. Our method preserves fine silhouette details and correctly defocuses through transparent or reflective surfaces. We demonstrate with these image regions of reflected vegetation which exhibit both detailed silhouettes and layered depth.}
    \label{fig:intro}
    \vspace{-0.3cm}
\end{figure*}

Shallow depth-of-field (DOF) is an aesthetically pleasing effect in photography. Defocus blur suppresses details in the foreground and background which are outside the DOF, while viewer attention is directed to the subject which is within the DOF. Large aperture lenses are necessary to produce shallow DOF with strong defocus blur in foreground and background regions. However, the ubiquitous smartphone cameras often have small apertures 1 to 2 mm in diameter. They are insufficient for producing strong defocus blur naturally, and can only produce sharper images with high DOF. For cameras with small aperture lenses, shallow DOF images are synthesized from sharp images in post-process.

The conventional approach to synthetic defocus blur is depth-based image blurring. The input is a color+depth (RGB-D) image where the depth image is acquired by depth sensors, recovered using structure from motion (SfM) or multi-view stereo (MVS) techniques~\cite{Schonberger_2016_CVPR,seitz2006comparison,6909904, Luo_2020_CVPR}, or estimated from the monocular color image itself~\cite{saxena2005learning,garg2016unsupervised}. The blur is weaker for pixels within the DOF and stronger outside the DOF. The spatially varying blur strength is called the defocus map. Alternatively, the defocus map can be estimated from any existing defocus blur in the image, then magnified to compress the original DOF~\cite{ZHUO20111852,bae2007defocus}. The defocus map can also be approximated from image segmentation such that the blurring is applied only to regions outside the subject's image segment. A hybrid approach applies depth-based blurring in context regions while keeping the subject's image segment in perfect focus~\cite{10.1145/3197517.3201329}. However, some image features remain challenging for this conventional approach due to limitations in depth estimation. For example, silhouette edges have sub-pixel depth features and suffer from depth estimation errors; transparent and reflective surfaces do not admit a single depth value, hence blur strength, per pixel~\cite{10.1145/3272127.3275032}.

Light field rendering is another approach for synthesizing shallow DOF images~\cite{10.1145/237170.237199}. The 4D light field can be acquired manually using a single lens --- the user captures images of the same scene from an array of viewpoints. However, the user has to laboriously scan multiple viewpoints within the simulated aperture using 2D hand motion~\cite{levoy2012synthcam}, while a simpler 1D trajectory only simulates an elongated synthetic aperture~\cite{5559009}. Viewpoint interpolation techniques can generate the dense light field from sparse acquisition viewpoints, but these viewpoints must align with predefined positions, necessitating a calibrated multi-lens arrangement~\cite{10.1145/2980179.2980251}. The novel learning-based approach estimates the dense light field from a single color image. However, it does not generalize well to image categories unseen during training~\cite{srinivasan2017learning}. Overall, light fields are powerful intermediate representations for generating defocus effects, but they are challenging to acquire.

We introduce a method for simulating defocus blur directly from a burst of sharp images taken with a small aperture camera, without explicit depth-based image blurring. Our method takes a short handheld burst of images as input. The images are first aligned and refocused conventionally at the focus depth. Our deep learning model then generates a shallow DOF output image, where the simulated aperture size equals the user's lateral hand translation during burst acquisition. In other words, the simulated defocus blur for each region is as strong as its disparity. Our method does not require multiple lenses, depth sensing capability, or laborious manual acquisition. Compared with conventional depth-based image blurring, our method successfully tackles challenging image regions such as transparent or reflective surfaces, as well as fine object silhouettes (Fig.~\ref{fig:intro}).




\section{Proposed Method}
\begin{figure*}
\centering
    \vspace{-0.6cm}
    \includegraphics[width=\linewidth]{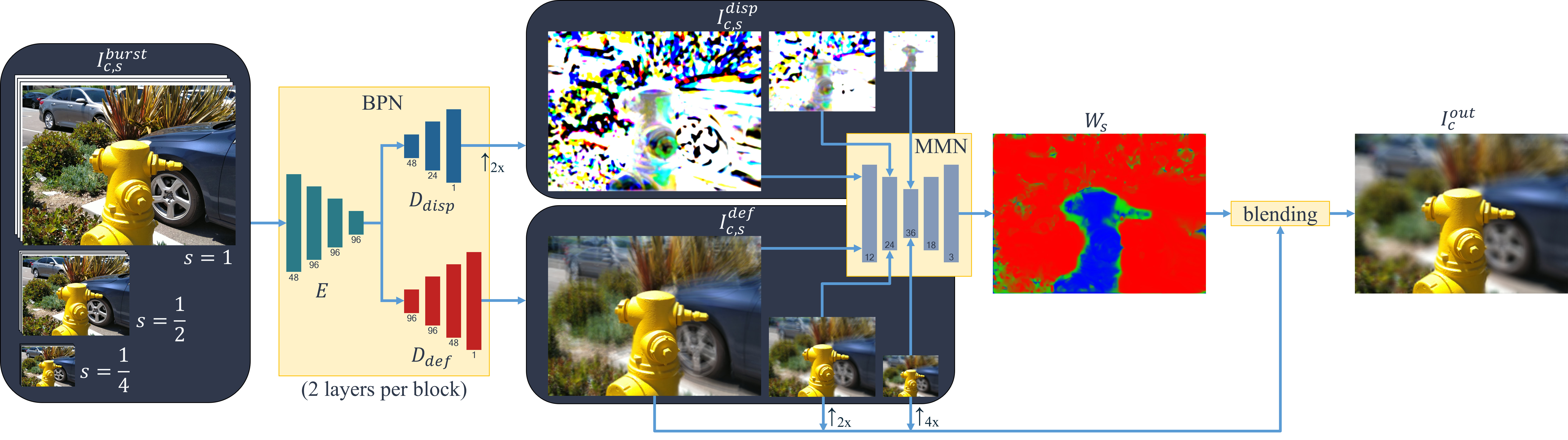}
	\vspace{-0.6cm}
    \caption{From a handheld burst, the Blur Prediction Network (BPN) infers defocus blur images $I^{def}_{c,s}$ and corresponding disparity maps $I^{disp}_{c,s}$ in parallel. The Multi-scale Merging Network (MMN) merges $I^{def}_{c,s}$ from different scales with blending weights $W_s$. In the absence of MMN, the full-scale BPN output $I^{def}_{c,s=1}$ shows significant ghosting artifacts in high disparity regions.}
    \vspace{-0.6cm}
    \label{figure:full_setup}
\end{figure*}

A handheld burst is captured for each scene. During burst acquisition, the camera moves laterally over a small distance that equals the diameter of the simulated aperture. We assume the trajectory is from top to bottom in the camera's reference frame. In general, any captured burst can be re-oriented to achieve a top-to-bottom trajectory. The trajectory need not be perfectly linear, as slight deviations are expected and leveraged by the algorithm. In the case of spontaneous hand motion over a short burst, the trajectory still approximates straight lines or gentle curves~\cite{10.1145/3306346.3323024,Park_2014_CVPR}. We assume the longitudinal translation is much less than scene depth, and the lateral translation is much greater than the diameter of the real aperture.

The burst undergoes two preprocessing steps using classical computer vision algorithms. First, the images are warped to align all camera orientations. This warping is a homography transformation calculated by aligning distant feature points or using onboard gyroscopes~\cite{6831799,Park_2014_CVPR}. Second, the warped images are refocused by shifting in image X-Y until the in-focus region is aligned across all frames. This is similar to auto focusing by contrast detection. For portraits, it is facilitated by face detection. The aligned and refocused burst of images is the input to our ML model (Fig.~\ref{figure:full_setup}).


The aligned and refocused images can be considered an incomplete light field $L(x,y,u,v)$ with only a subset of viewpoints $(u,v)$ available. However, we don't perform camera localization and therefore don't know the exact $(u,v)$ coordinates. We do assume the image sequence to have decreasing $v$ coordinates due to the top-to-bottom trajectory. Fig.~\ref{fig:49views} (e) shows an example viewpoint trajectory for a 9-frame handheld burst, represented as an incomplete light field.

The blur prediction network (BPN) is employed on one color channel at a time. BPN is a U-Net~\cite{RFB15a} that directly predicts the defocus blur image for each color channel. The 3 output channels combine to produce the full RGB output. In addition to the defocus decoder $D_{def}$, the U-Net has a second decoder $D_{disp}$ whose purpose is twofold -- First, it generates an estimated disparity output to facilitate \emph{multi-task training} (Sec. \ref{sec:multitask}). Second, the additional disparity output infers a reliability measure for merging BPN results during \emph{multi-scale inference} (Sec. \ref{sec:multiscale}).


\subsection{Multi-task Training}
\label{sec:multitask}
The second decoder $D_{disp}$ is appended to the BPN bottleneck layer. $D_{disp}$ is tasked with estimating the scene disparity, and it is supervised by the ground truth disparity during training. This supervision ensures that the encoder $E$ is attentive to the disparity across the input frames. However, the disparity output is not used to generate defocus blur --- all defocus blur is generated in parallel with the estimated disparity.

\subsection{Multi-scale Inference}
\label{sec:multiscale}
In some scenes, image regions can have disparities high enough to exceed the receptive field of BPN, then it becomes impossible for BPN to observe the disparity and to output the correct defocus blur image. This is caused by the foreground or background context being too far from the plane of focus, or excessive camera translation during burst acquisition. Deepening the BPN U-Net or using larger convolutional filters can expand the receptive field to cover larger disparities, but we choose a multi-scale approach that is more efficient.

First, BPN is employed on inputs $I^{burst}_{c,s}$ at original, half, and quarter scales $s$, producing both defocus blur image $I^{def}_{c,s}$ and disparity $I^{disp}_{c,s}$ at each of the corresponding scales. Then, the multi-scale merging network (MMN) predicts the per-pixel blending weight $W_s$. The merged output is thus a weighted sum of BPN's outputs: $I^{out}_c=\sum_s W_s\times I^{def}_{c,s}$.

MMN predicts blending weights $W_s$ from the magnitude and consistency of disparity maps ---  high or inconsistent disparity across color channels indicate the scene disparity has reached or exceeded BPN's receptive field, and $I^{def}_{c,s}$ at a lower scale should be given more weight. Low disparity indicates the scene region is sharp, and $I^{def}_{c,s}$ at a higher scale should be given more weight.

\section{Experiments}
We train the BPN and MMN models separately, as BPN by itself should correctly defocus images if the scene disparity were not excessive. Therefore, BPN is trained first, then its parameters are frozen while MMN is trained second within the full pipeline (Fig.~\ref{figure:full_setup}).

\begin{figure}
\centering
    \includegraphics[width=\linewidth]{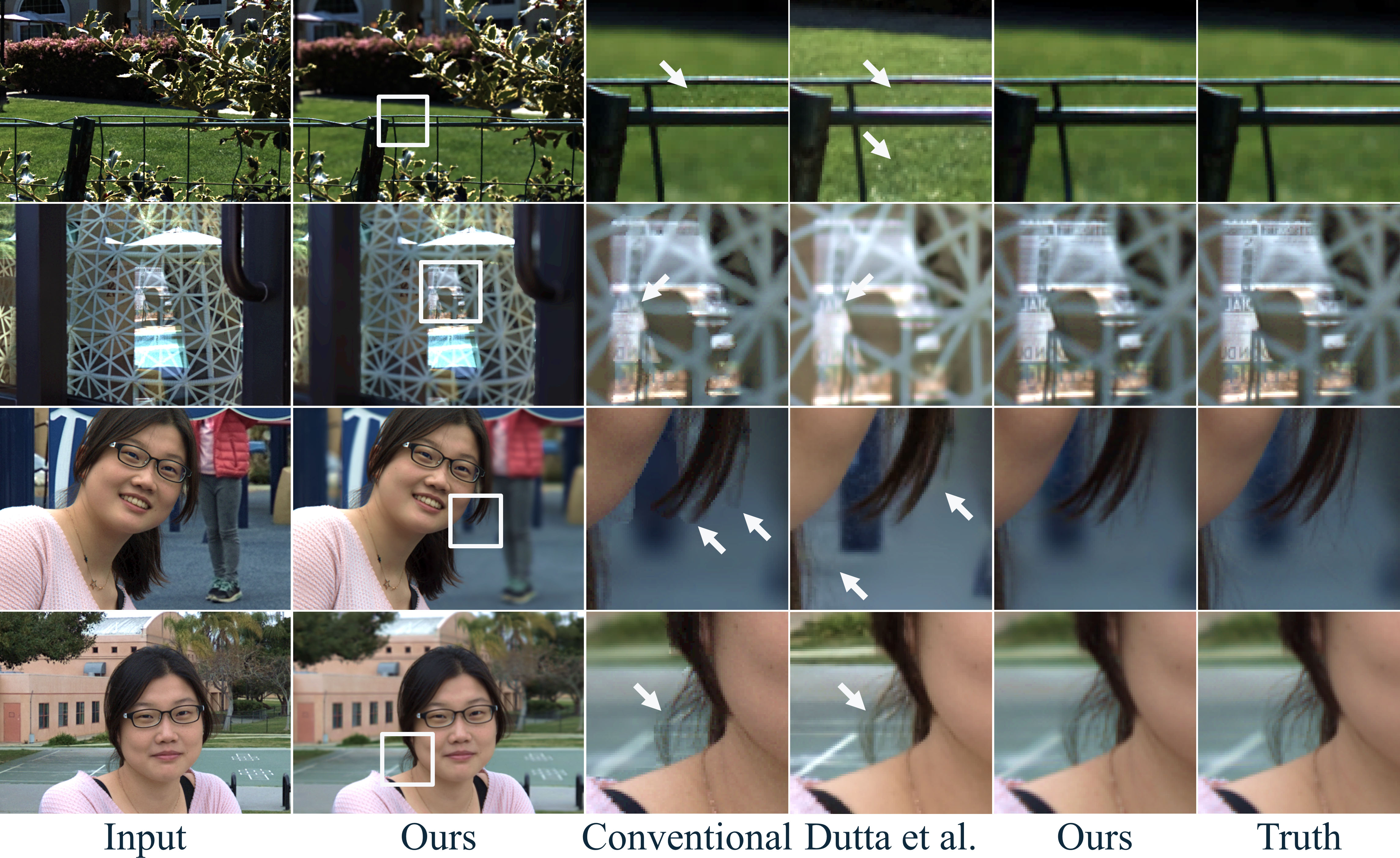}
    \vspace{-0.7cm}
    \caption{Our method defocuses correctly around detailed silhouettes and within areas of layered depth (top rows). Our method is well-suited to portrait photography (bottom rows).}
    \label{fig:visual}
    \vspace{-0.6cm}
\end{figure}

Light field datasets are leveraged for model training and evaluation --- light field datasets support straightforward simulation of ground truth as well as burst inputs for supervised learning. Specifically, both models are trained using the DeepFocus synthetic light field dataset~\cite{10.1145/3272127.3275032}. This dataset consists of rendered 3D scenes of various 3D models, posed randomly, and colored by random textures. For quantitative evaluation, our pipeline is employed on the Stanford Lytro Light Field dataset~\cite{raj2016stanford}, a collection of natural scenes acquired using the Lytro Illum camera. We also acquire Lytro Illum images of certain challenging scenes in order to highlight the improvement over conventional algorithms (Fig.~\ref{fig:visual}).

Additionally, we capture real handheld burst images using a smartphone camera (Fig.~\ref{fig:intro}, ~\ref{figure:full_setup}, and~\ref{fig:additional}), and we also extract consecutive frames from computer game footage (Fig.~\ref{fig:additional}, right). These images are only for visual evaluation as they lack ground truths for quantitative evaluation.

\subsection{Training Setup}

\begin{figure}
    \centering
    \begin{subfigure}[b]{0.19\linewidth}
        \centering
        \includegraphics[width=\textwidth]{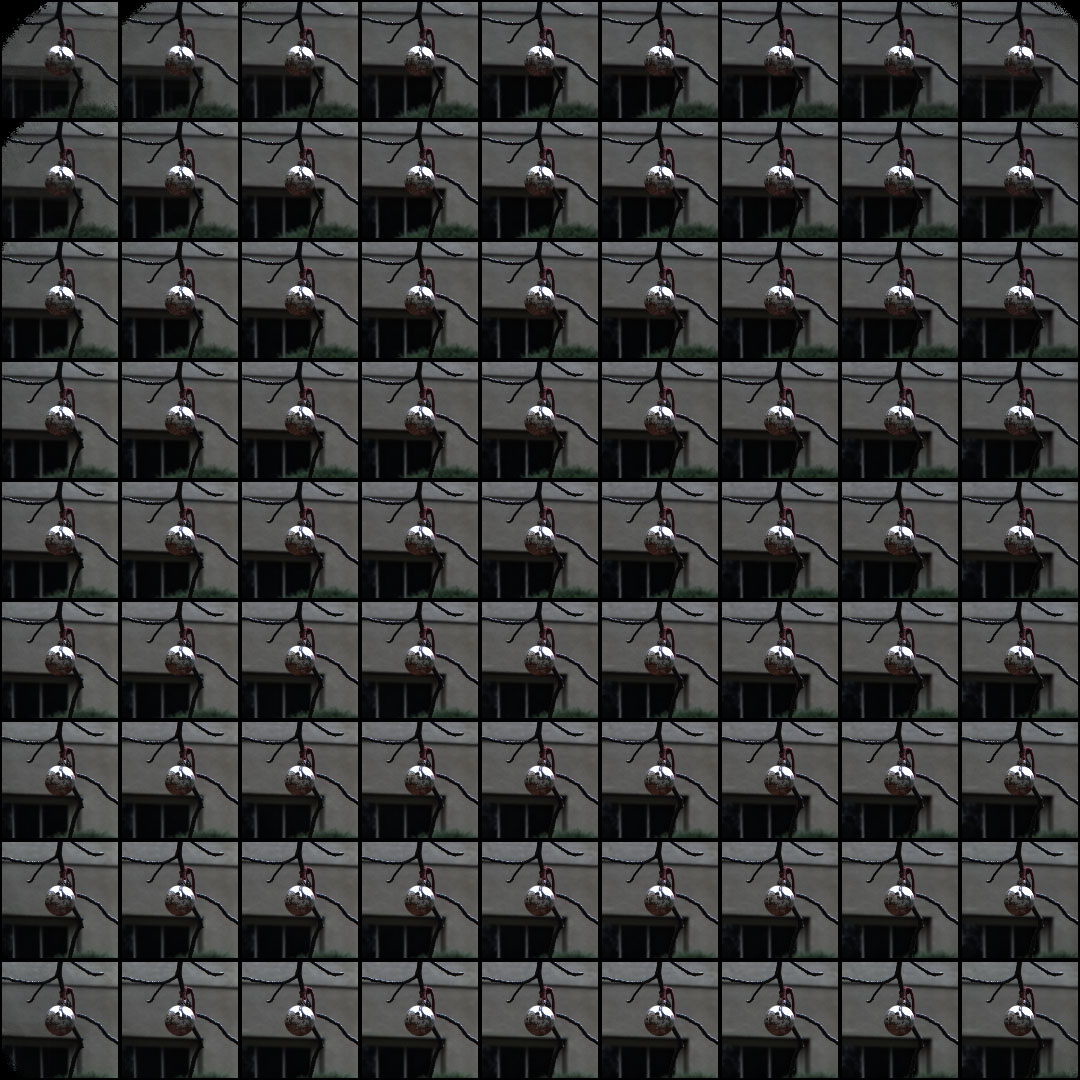}
        \caption{}
    \end{subfigure}
    \hfill
    \begin{subfigure}[b]{0.19\linewidth}
        \centering
        \includegraphics[width=\textwidth]{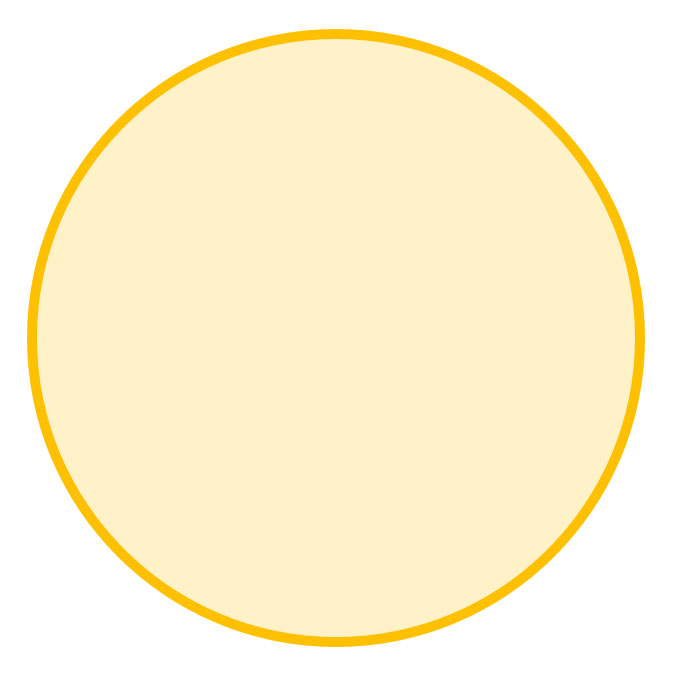}
        \caption{}
    \end{subfigure}
    \hfill
    \begin{subfigure}[b]{0.19\linewidth}
        \centering
        \includegraphics[width=\textwidth]{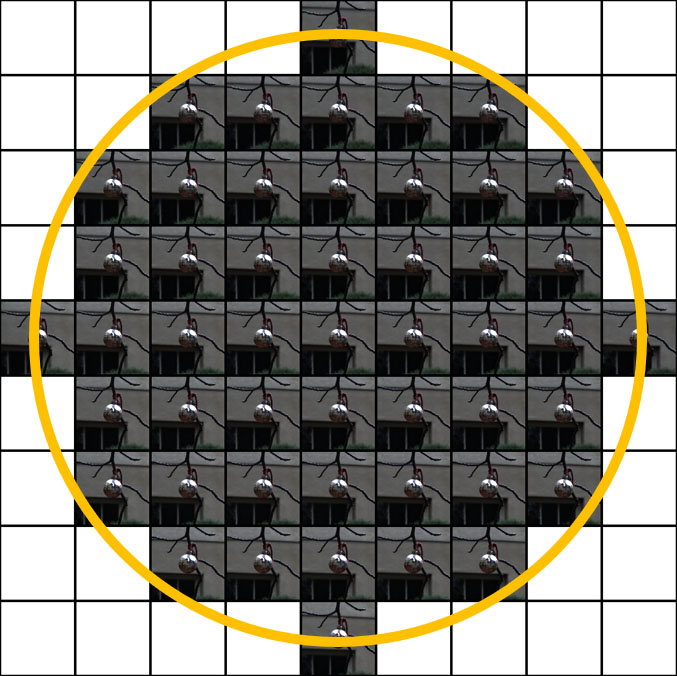}
        \caption{}
    \end{subfigure}
    \hfill
    \begin{subfigure}[b]{0.19\linewidth}
        \centering
        \includegraphics[width=\textwidth]{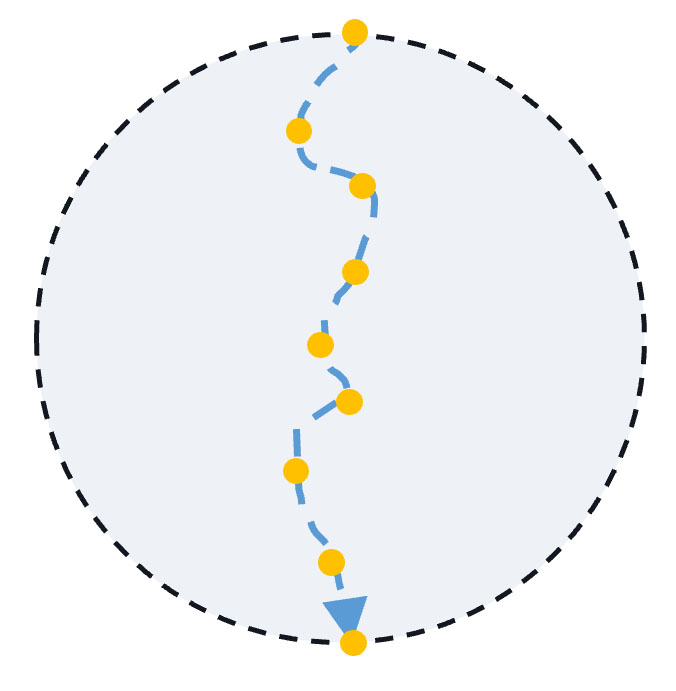}
        \caption{}
    \end{subfigure}
    \hfill
    \begin{subfigure}[b]{0.19\linewidth}
        \centering
        \includegraphics[width=\textwidth]{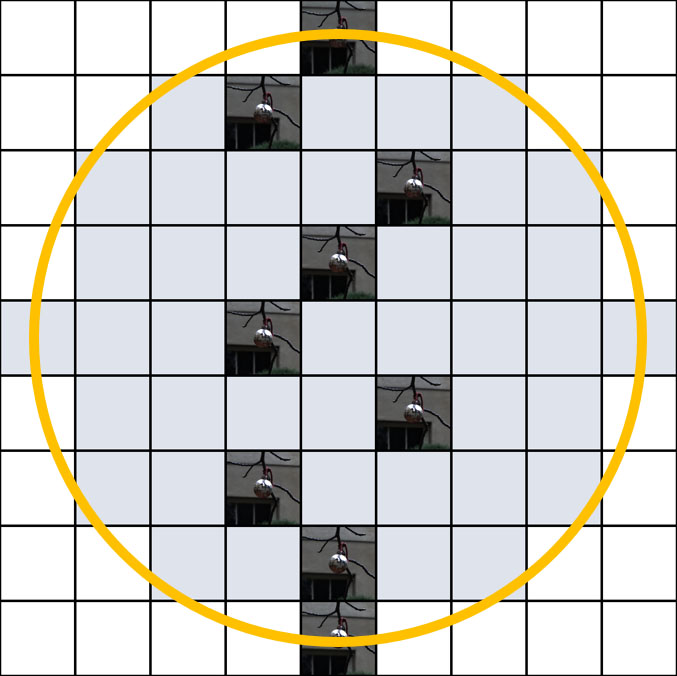}
        \caption{}
    \end{subfigure}
    \vspace{-0.3cm}
    \caption{a) A dense light field consists of $9\times9$ sub-aperture images. b) The simulated aperture. c) The 49 sub-aperture images that average to the simulated aperture ground truth. d) Camera viewpoints during handheld burst acquisition. e) The 9 sub-aperture images that simulate the handheld burst.}
    \label{fig:49views}
    \vspace{-0.3cm}
\end{figure}

The input to our pipeline is a sequence of light field sub-aperture images corresponding to the viewpoint trajectory of a simulated top-to-bottom handheld burst. The trajectory deviates from a perfect line with randomly selected viewpoints slightly offset to the left or right. The ground truth for both BPN and MMN supervision is produced by averaging all sub-aperture images whose viewpoints lie within a circle representing the shape of the circular simulated aperture. Within the light field consisting of $9\times9=81$ sub-aperture viewpoints, the circular subset covers $49$ viewpoints (Fig.~\ref{fig:49views}). 




Specifically, the ground truth $I^{gt}_c$ is calculated from the parameterized 4D light field $L_c(x,y,u,v)$ where $(x,y)$ denotes the 2D ray direction, and $(u,v)$ denotes the sub-aperture viewpoint coordinate. This parameterization maps straightforwardly to burst acquisition where $(x,y)$ is pixel coordinate within a frame, and $(u,v)$ denotes its camera viewpoint. With $x,y,u,v$ all in discrete pixel and viewpoint coordinates:
\begin{equation}
    I^{gt}_c(x,y) = \frac{1}{|A|}\sum_{(u,v)\in A}^{} L_c(x-\alpha u, y-\alpha v, u, v)
    \label{eq:photography_op}
\end{equation}
where A is the subset of viewpoints making up the simulated aperture (Fig.~\ref{fig:49views} (c)), and $\alpha$ is the refocusing factor. The input images are similarly refocused by $(x,y)\leftarrow(x-\alpha u, y-\alpha v)$. We vary $\alpha$ between $0$ and $4$ during training as data augmentation but keep $\alpha=0$ for inference. $I^{disp\_gt}$ is lifted directly from the dataset then biased according to the $\alpha$ setting. Other augmentations include color inversion and image scaling.

BPN training is supervised through this loss function:
\begin{equation} \label{eq:bnp_loss}
\begin{split}
\L_{bpn} = & \L_{ssim}(I^{def}_{c,1}, I^{gt}_c) + \lambda_c \lVert I^{def}_{c,1} - I^{gt}_c \rVert_{1} \\ + & \lambda_d \lVert I^{disp}_{c,1} - I^{disp\_gt}_c \rVert_{1}
\end{split}
\end{equation}
where $\L_{ssim}(x,y)$ is negative multiscale structural similarity~\cite{1292216} with two scales weighted $[0.9, 0.1]$ and window size $7$; the weights $\lambda_c$ and $\lambda_d$ are experimentally set at $0.5$ and $0.1$. The MMN loss function is defined as:
\begin{equation} \label{eq:msn_loss}
\L_{mmn} = \L_{ssim}(I^{out}_c, I^{gt}_c)
\end{equation}
where $\L_{ssim}$ uses the same parameters, but is applied between the merged image $I^{out}_c$ and the ground truth $I^{gt}_c$.

\subsection{Evaluation}
\textbf{Qualitative} Fig.~\ref{fig:visual} shows a direct visual comparison with the conventional method based on defocus maps, here represented by the Lens Blur feature in the Adobe Photoshop application. The direct monocular to DOF method of Dutta et al.~\cite{Dutta_2021_CVPR} is also shown for comparison. Each scene is acquired by the Lytro Illum camera, whose light field images simulate handheld bursts for our method according to Fig.~\ref{fig:49views} (e), as well as provide RGB-D images for the conventional method.

\begin{figure}[h]
\centering
    \vspace{-0.2cm}
    \includegraphics[width=\linewidth]{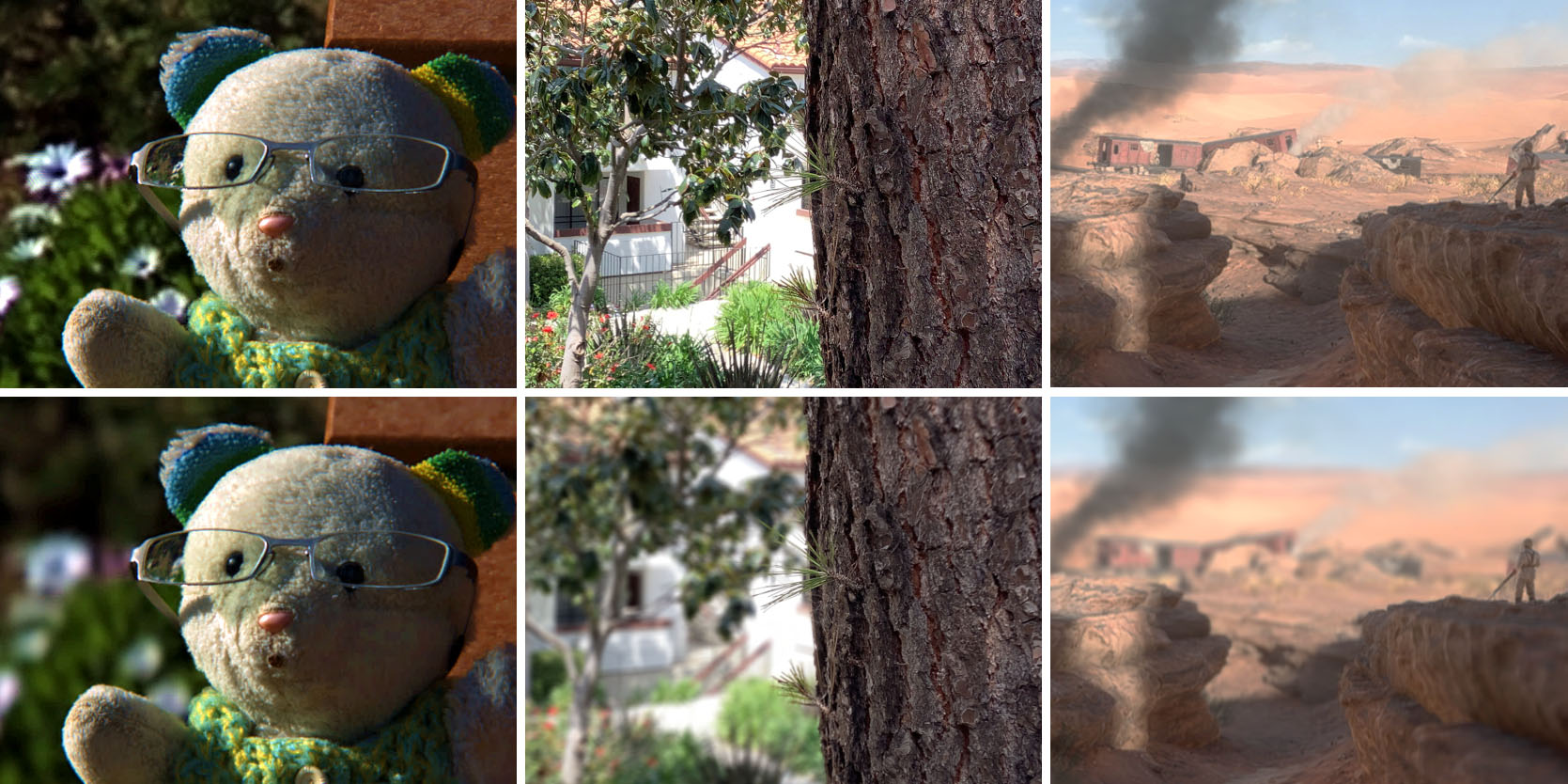}
	\vspace{-0.6cm}
	\caption{Our method generalizes well to diverse natural and synthetic scenes. Top row: one of the input frames. Bottom row: our results. Left to right: Lytro Illum acquisition, handheld burst acquisition, computer game sequence.}
	\label{fig:additional}
	\vspace{-0.7cm}
\end{figure}

Our method avoids artifacts due to limitations in the depth-based defocus map. Particularly, our method successfully defocuses sub-pixel features and pixels with more than one depth value. Common portrait artifacts occurring near loose hairs are significantly reduced with our method. Fig.~\ref{fig:additional} shows additional results in other natural and synthetic scenes.

\begin{table}[b]
\vspace{-4mm}
\centering
\caption{Performance on the Stanford Lytro Light Field dataset. "Center view only" is provided as a baseline.}
\vspace{-0.3cm}
\label{table:quantitative}

\begin{tabular}{l c c} 
\toprule
{} & {SSIM$\uparrow$} & {LPIPS$\downarrow$} \\

\midrule
{Kalantari et al. (4 corner viewpoints)} & {0.941} & {0.0580} \\
{Ours (4-frame burst)} & {\textbf{0.972}} & {\textbf{0.0491}} \\
\midrule
{Ours (9-frame burst)} & {0.980} & {0.0373} \\
{Center view only, no defocus} & {0.894} & {0.1425} \\
\bottomrule
\end{tabular}


\end{table}

\textbf{Quantitative} The light field rendering technique synthesizes defocus effects free from the limitations of conventional defocus maps. Our method is compared with the viewpoint interpolation technique of Kalantari et al.~\cite{10.1145/2980179.2980251}, which reconstructs dense light fields from only 4 corner viewpoints.

The Stanford Lytro Light Field Archive~\cite{raj2016stanford} is leveraged to simulate burst images for our algorithm according to Fig.~\ref{fig:49views} (e); it also provides the 4-viewpoint sub-aperture images for the prior work's viewpoint interpolation algorithm. We choose random handheld trajectories where the center of mass of the viewpoints coincides with the center of the full aperture; we also set our inputs to 4-frame bursts to align with the 4 corner viewpoint requirement of the prior work. The prior work interpolates $8\times8$ viewpoints instead of $9\times9$, missing the topmost and left most viewpoints in Fig.\ref{fig:49views} (c). Therefore, these 2 viewpoints are substituted by sub-aperture images directly lifted from the dataset. Eq.~\ref{eq:photography_op} is used to produce the defocus blur images from prior work's interpolated viewpoints; it is also used to produce truth images from the dataset's raw light fields. Image quality is measured in SSIM~\cite{wang2004image} and LPIPS~\cite{Zhang_2018_CVPR}. The results are shown in Table~\ref{table:quantitative}.

Our method produces superior results to the prior work algorithm at the same number of input frames. Notably, our algorithm only requires loosely structured handheld bursts from a single camera, while the prior work algorithm requires multiple cameras at exactly arranged viewpoints. Furthermore, our algorithm scales well with additional input frames, thus it can take advantage of increased burst acquisition frame rates in real-world camera systems.

Our algorithm is efficient. On a computer with Intel Core i7-8700K and Nvidia RTX 2080, our algorithm processes a benchmark scene with 4-frame input in 1.97s, while the prior work requires 487s to interpolate all 49 needed viewpoints.

\section{Conclusion}
We address the problem of simulating shallow DOF by proposing a novel approach that produces defocus blur directly from observable scene disparity in a handheld burst. Our method does not require any depth sensing capabilities, it leverages existing image warping and alignment infrastructure, and it is agnostic to image content.

By eschewing the defocus map used in the conventional approach, our method succeeds in image regions that are otherwise challenging, such as transparencies, reflections, and fine silhouette details.

Compared with existing light field acquisition, completion, and rendering techniques, our method produces superior results yet requires simple monocular burst imaging instead of multi-lens setups.

Our method generalizes well to diverse scenes. Indeed, our model is trained on a fully synthetic dataset, yet produces satisfactory results on natural and synthetic scenes alike.
\clearpage


%


\bibliographystyle{IEEEbib}
\bibliography{strings,bib}

\end{document}